\documentclass[aps,pre,superscriptaddress,twocolumn,nofootinbib]{revtex4-2}
\usepackage[utf8]{inputenc}
\usepackage[T1]{fontenc}
\usepackage{tabularx,multirow,booktabs,blindtext}

\usepackage{multirow} 
\usepackage{graphicx} 
\usepackage[dvipsnames]{xcolor}
\usepackage{amsmath,amssymb} 
\usepackage{subcaption}

\graphicspath{ {./}{./images/} }
\newcommand{\ourdata}{ZooLake } 

\newcommand{\pb}[1]{\textbf{\textcolor{green}{#1}}}
\newcommand{\C}{\vec C}
\newcommand{\s}{\vec \sigma}
\DeclareMathOperator\erf{erf}

\begin{document}
\title[Vision transformers]{Ensembles of Vision Transformers as a New Paradigm for Automated Classification in Ecology
} 
\author{S. Kyathanahally}
\email{sreenath.kyathanahally@eawag.ch}
\affiliation{Eawag, \"Uberlandstrasse 133, CH-8600 D\"ubendorf, Switzerland}
\author{T. Hardeman}
\affiliation{Eawag, \"Uberlandstrasse 133, CH-8600 D\"ubendorf, Switzerland}
\author{M. Reyes}
\affiliation{Eawag, \"Uberlandstrasse 133, CH-8600 D\"ubendorf, Switzerland}
\author{E. Merz}
\affiliation{Eawag, \"Uberlandstrasse 133, CH-8600 D\"ubendorf, Switzerland}
\author{T. Bulas}
\affiliation{Eawag, \"Uberlandstrasse 133, CH-8600 D\"ubendorf, Switzerland}
\author{P. Brun}
\affiliation{WSL, Z\"urcherstrasse 111, CH-8903 Birmensdorf, Switzerland}
\author{F. Pomati}
\affiliation{Eawag, \"Uberlandstrasse 133, CH-8600 D\"ubendorf, Switzerland}
\author{M. Baity-Jesi}
\email{marco.baityjesi@eawag.ch}
\affiliation{Eawag, \"Uberlandstrasse 133, CH-8600 D\"ubendorf, Switzerland}

\date{\today}

\begin{abstract}
Monitoring biodiversity is paramount to manage and protect natural resources. Collecting images of organisms over large temporal or spatial scales is a promising practice to monitor the biodiversity of natural ecosystems, providing large amounts of data with minimal interference with the environment. Deep learning models are currently used to automate classification of organisms into taxonomic units. However, imprecision in these classifiers introduces a measurement noise that is difficult to control and can significantly hinder the analysis and interpretation of data. {We overcome this limitation through ensembles of Data-efficient image Transformers (DeiTs), which not only are easy to train and implement, but also significantly outperform} the previous state of the art (SOTA). We validate our results on ten ecological imaging datasets of diverse origin, ranging from plankton to birds. On all the datasets, we achieve a new SOTA, with a reduction of the error with respect to the previous SOTA ranging from {29.35\%} to {100.00\%}, and often achieving performances very close to perfect classification. 
Ensembles of DeiTs perform better not because of superior single-model performances but rather due to smaller overlaps in the predictions by independent models and lower top-1 probabilities. This increases the benefit of ensembling, especially when using geometric averages to combine individual learners. While we only test our approach on biodiversity image datasets, our approach is generic and can be applied to any kind of images.
\end{abstract}

\maketitle

\section{Introduction}
Biodiversity monitoring is critical because it serves as a foundation for assessing ecosystem integrity, disturbance responses, and the effectiveness of conservation and recovery efforts~\cite{kremen:94,jetz:19, kuhl:20}. Traditionally, biodiversity monitoring relied {on empirical data collected manually}~\cite{witmer:05}. This is time-consuming, labor-intensive, and costly. Moreover, such data can contain sampling biases as a result of difficulties controlling for observer subjectivity and animals' responses to observer presence~\cite{mcevoy:16}. 
These constraints severely limit our ability to estimate the abundance of natural populations and community diversity, reducing our ability to interpret their dynamics and interactions.
Counting wildlife by humans has a tendency to greatly underestimate the number of individuals present~\cite{hodgson:18, tuia:22}. Furthermore, population estimates based on extrapolation from a small number of point counts are subject to substantial uncertainties and may fail to represent the spatio-temporal variation in ecological interactions (e.g. predator-prey), leading to incorrect predictions or extrapolations~\cite{soranno:14, tuia:22}. While human-based data collection has a long history in providing the foundation for much of our knowledge of where and why animals dwell and how they interact, present difficulties in wildlife ecology and conservation are revealing the limitations of traditional monitoring methods~\cite{tuia:22}.

Recent improvements in imaging technology have dramatically increased the data-gathering capacity by lowering costs and widening the scope and coverage compared to traditional approaches, opening up new paths for large-scale ecological studies~\cite{tuia:22}. Many formerly inaccessible places of conservation interest may now be examined by using high-resolution remote sensing~\cite{luque:18}, and digital technologies such as camera traps~\cite{burton:15,rowcliffe:08,steenweg:17} are collecting vast volumes of data non-invasively. Camera traps are low-cost, simple to set up, and provide high-resolution image sequences of the species that set them off, allowing researchers to identify the animal species, their behavior, and interactions including predator-prey, competition and facilitation. Several cameras have already been used to monitor biodiversity around the world, including underwater systems~\cite{orenstein:20,merz:21}, making camera traps one of the most widely-used sensors~\cite{steenweg:17}.
In biodiversity conservation initiatives, camera trap imaging is quickly becoming the gold standard~\cite{burton:15,rowcliffe:08}, as it enables for unparalleled precision monitoring across enormous expanses of land. 

However, people find it challenging to analyze the massive amounts of data provided by these devices. The enormous volume of image data generated by modern gathering technologies for ecological studies is too large to be processed and analyzed at scale to derive compelling ecological conclusions~\cite{farley:18}. Although online crowd-sourcing platforms could be used to annotate images~\cite{jamison:15}, such systems are unsustainable due to the exponential expansion in data acquisition and to the insufficient expert knowledge that is most often required for the annotation. In other words, we need tools that can automatically extract relevant information from the data and help to reliably understand how ecological processes act across space and time.

Machine learning has proven to be a suitable methodology to unravel the ecological insights from massive amounts of data~\cite{kwok:19}. Detection and counting pipelines have evolved from imprecise extrapolations from manual counts to machine learning-based systems with high detection rates~\cite{norouzzadeh:18,willi:19,tabak:19}. Using deep learning (DL) to detect and classify species for the purpose of population estimation is becoming increasingly common~\cite{henrichs:21,norouzzadeh:18,willi:19,tabak:19,kyathanahally:21,py:16,dai:17,Lee:16,luo:18,islam:20}. DL models, most often with convolutional neural network (CNN) like architectures~\cite{kyathanahally:21,willi:19,tabak:19,dai:17,luo:18,norouzzadeh:18}, have been the standard thus far in biodiversity monitoring. Although these models have an acceptable performance, they often unreliably detect minority classes~\cite{kyathanahally:21}, require 
{a very well-tailored model selection and training,}
large amounts of data~\cite{tabak:19}, and have a non-negligible error rate that negatively influences the modeling and interpretation of the outcoming data. 
Thereupon, it is argued that {many DL}-based monitoring systems cannot be deployed in a fully autonomous way if one wants to ensure a reliable-enough classification~\cite{green:20,schneider:20}.

Recently, following their success in natural language processing applications~\cite{vaswani:17}, transformer architectures were adapted to computer vision applications. The resulting structures, known as vision transformers~(ViTs)~\cite{dosovitskiy:20}, differ from CNN-based models\pb{,} that use image pixels as units of information, in using image patches, and employing an attention mechanism to weigh the importance of each part of the input data differently.
Vision transformers have demonstrated encouraging results in several computer vision tasks, outperforming the state of the art (SOTA) in several paradigmatic datasets, and paving the way for new research areas within the branch of deep learning. 

In this article, we use a specific kind of ViTs, Data efficient image Transformers (DeiTs)~\cite{touvron:20},
for the classification of biodiversity images such as plankton, coral reefs, insects, birds and large animals (though our approach can also be applied in different domains). 
We show that while the single-model performance of DeiTs matches that of alternative approaches, ensembles of DeiTs {(EDeiTs)} significantly outperform the previous SOTA, both in terms of higher accuracy and of better classification of minority classes (\textit{i.e.} rare species). 
We see that this mainly happens because of a higher disagreement in the predictions, with respect to other model classes, between independent DeiT models. 
{Finally, we find that while CNN and ViT ensembles perform best when individual learners are combined through a sum rule, EDeiTs perform best when using a product rule.}

\section{Results}
\label{sec:result}
\begin{figure*}[tb!]
\centering
\includegraphics[width=\textwidth]{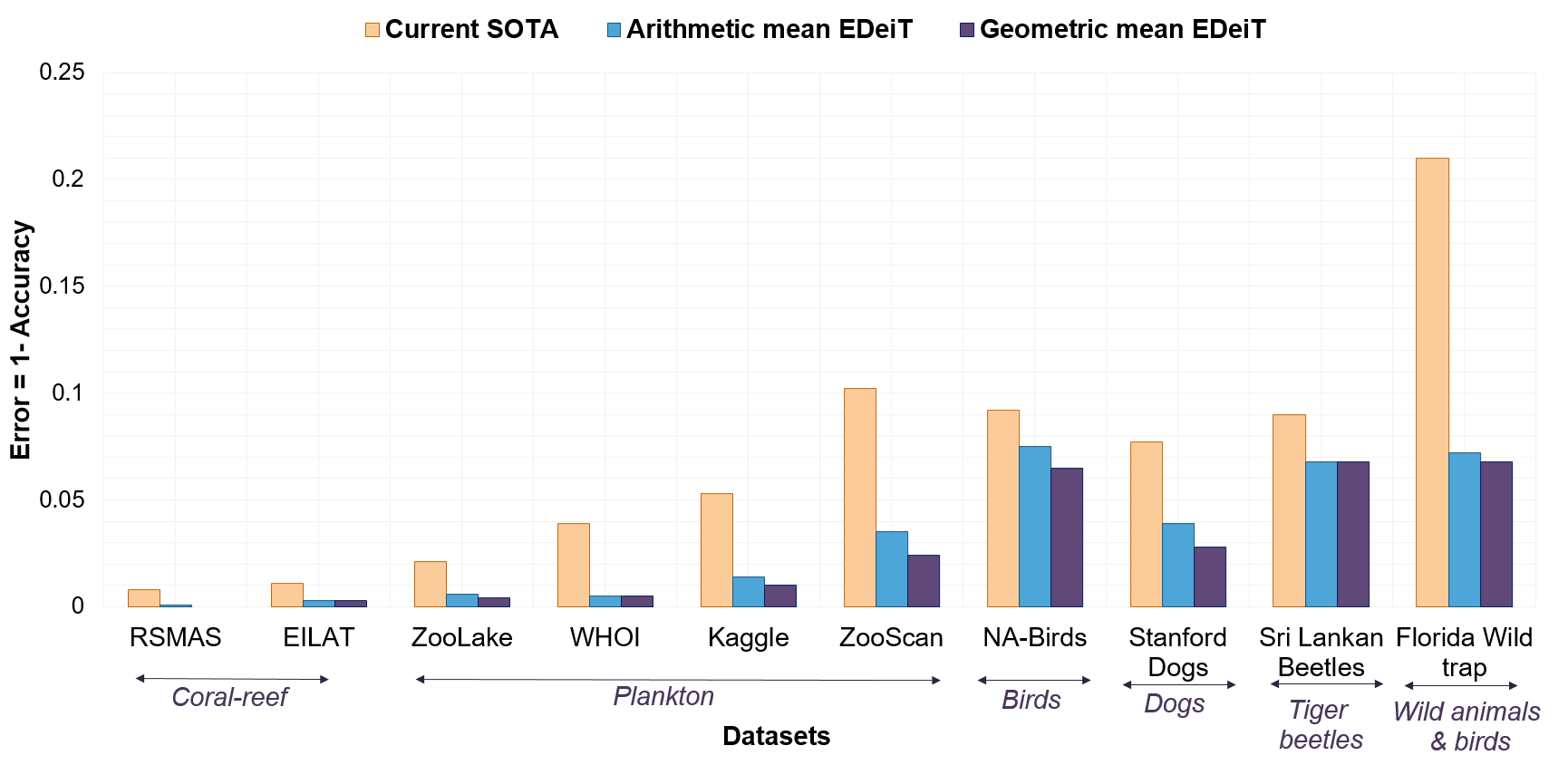}
\caption{Comparing EDeiTs to the previous SOTA. For each dataset, we show the error, which is the fraction of misclassified test images ($1-\text{accuracy}$). The error of the existing SOTA model is shown in orange. For the ensembles of DeiTs, we show two ways of combining the individual learnings: through arithmetic (blue) and geometric (purple) averaging. The purple bar for RSMAS is absent because \textit{all} the test examples were classified correctly by the EDeit with geometric averaging.
Independent of the ensembling rule, our models outperform current SOTA models on a consistent basis.
}
\label{fig:comparison}
\end{figure*}

\subsection{A new state of the art}
We trained EDeiTs on several ecological datasets, spanning from microorganisms to large animals, including images in color as well as in black-and-white, with and without background; and including datasets of diverse sizes and with varying numbers of classes, both balanced and unbalanced.

Details on the datasets are provided in Sec.~\ref{sec:data}.
As shown in Fig.~\ref{fig:comparison}, the error rate of EDeiTs is drastically smaller than the previous SOTA, across all datasets.
In App.~\ref{app:comparisons-sota} we provide a detailed comparison between our models' accuracy and F1-score and that of the previous SOTA.
Details on models and training are provided in Secs.~\ref{sec:models},~\ref{sec:implementation} and~\ref{sec:ensembling}.

\subsection{Individual models comparison}
We now show that the better performance of EDeiTs is not a property of the single models, but that it rather stems from the ensembling step.
To do this, we focus on the ZooLake dataset where the previous state of the art is an ensemble of CNN models~\cite{kyathanahally:21} that consisted of EfficientNet, MobileNet and DenseNet architectures.
In Tab.~\ref{tab:individual-performance}, we show the single-model performances of these architectures, and those of the DeiT-Base model (Sec.~\ref{sec:implementation}), which is the one we used for the results in Fig.~\ref{fig:comparison}. 
The accuracies and (macro-averaged) F1-scores of the two families of models (CNN and DeiT) when compared individually are in a similar range: the accuracies are between 0.96 and 0.97, and the F1-scores between 0.86 and 0.90.

\begin{table*}[t]
\centering
\caption{Summary of the performance of the individual models on the ZooLake dataset. The ensemble score on the rightmost column is obtained by averaging across either 3 or 4 different initial conditions.
The Best\_6\_avg model is an ensemble of DenseNet121, EfficientNet-B2, EfficientNet-B5, EfficientNet-B6, EfficientNet-B7 and MobileNet (combining learners through an arithmetic mean) models~\cite{kyathanahally:21}. The numbers in parentheses are the standard errors, referred to the last significant digit.}
\label{tab:individual-performance}
\begin{tabular}{|l|l|l|l|l|l|}
\hline
\textbf{Model} & \textbf{\begin{tabular}[c]{@{}l@{}}No. of \\ params \\ for each \\ model\end{tabular}} & \textbf{\begin{tabular}[c]{@{}l@{}}Accuracy \\ Mean\end{tabular}} & \textbf{\begin{tabular}[c]{@{}l@{}}F1-score \\ Mean\end{tabular}} & \textbf{\begin{tabular}[c]{@{}l@{}}Arithmetic \\ Ensemble\\ (accuracy/\\ F1-score)\end{tabular}} & \textbf{\begin{tabular}[c]{@{}l@{}}Geometric \\ Ensemble\\ (accuracy/\\ F1-score)\end{tabular}} \\ \hline
\textbf{Dense121} & 8.1M & 0.965(3) & 0.86(1) & 0.976/0.916 & 0.977/0.917\\ \hline
\textbf{Efficient-B2} & 9.2M & 0.9670(4) & 0.894(2) & 0.975/0.915 & 0.975/0.914\\ \hline
\textbf{Efficient-B5} & 30.6M & 0.964(2) & 0.87(1) & 0.971/0.891 & 0.971/0.898\\ \hline
\textbf{Efficient-B6} & 43.3M & 0.965(1) & 0.880(7) & 0.971/0.904 & 0.972/0.906\\ \hline
\textbf{Efficient-B7} & 66.0M & 0.968(1) & 0.893(4) & 0.974/0.913 & 0.974/0.920\\ \hline
\textbf{Mobile-V2} & 3.5M & 0.961(2) & 0.881(5) & 0.971/0.907 & 0.973/0.909\\ \hline
\textbf{Best\_6\_avg} & \begin{tabular}[c]{@{}l@{}} - \end{tabular} & - & - & 0.978/0.924 & 0.977/0.923\\\hline
\textbf{DeiT-Base} & 85.8M & 0.962(3) & 0.899(2) & 0.994/0.973 & 0.996/0.984\\ \hline
\end{tabular}
\end{table*}

\subsection{Ensemble comparison}
We train each of the CNNs in Tab.~\ref{tab:individual-performance} four times (as described in Ref.~\cite{kyathanahally:21}), with different realisations of the initial conditions, and show their arithmetic average ensemble and geometric average ensemble (Sec.~\ref{sec:ensembling}) in the last two columns. We also show the performance of the ensemble model developed in Ref.~\cite{kyathanahally:21}, which ensembles over the six shown CNN architectures.
We compare those with the ensembled DeiT-Base model, obtained through arithmetic average ensemble and geometric average ensemble over three different initial conditions of the model weights.

As can be expected, upon ensembling the individual model performance improves sensibly. However, the improvement is not the same across all models. While all ensembled CNNs perform similarly to each other (with $F1-\text{score}\leq0.924$), ensembled DeiTs reach an almost perfect classification accuracy (with the F1-score reaching $0.983$). With arithmetic average ensembling, we have 15 misclassifications out of 2691 test images, 14 of which on {classes not associated to a specific taxon} classes (see App.~\ref{app:zoolake}). With geometric averaging, the performance is even higher {with only 10 out of 2691 misclassified test images}.

\subsection{Why DeiT models ensemble better}
To understand the better performance of DeiTs upon ensembling, we compare CNNs with DeiTs when ensembling over three models. For CNNs, we take the best EfficientNet-B7, MobileNet and Dense121 models from Ref.~\cite{kyathanahally:21} (each had the best validation performance from 4 independent runs). For DeiTs, we train a DeiT-Base model three times (with different initial weight configurations) and ensemble over those three.

\begin{figure*}[tbh!]
    \centering
    \includegraphics[width=0.99\linewidth]{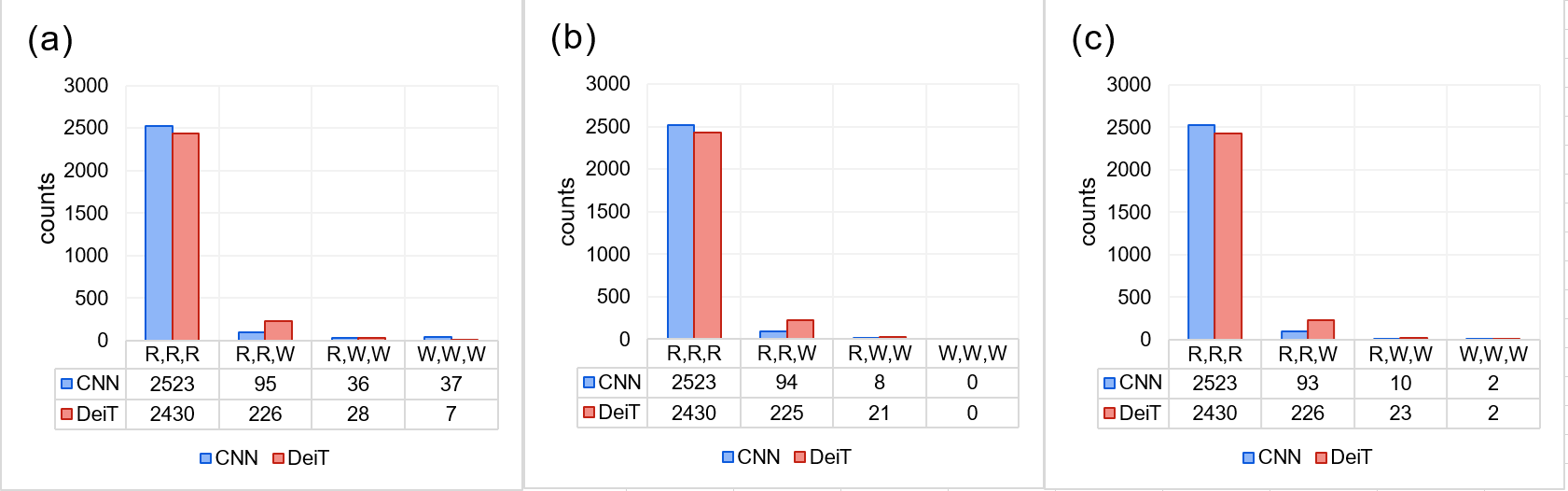}
    \caption{
    Comparison between three-model ensemble models based on CNNs and on DeiTs on the ZooLake test set. The bar heights indicate how often each combination (RRR, RRW, RWW, WWW) appeared. RRR indicates that all the models gave the right answer, RRW means that one model gave a wrong answer, and so on. The numbers below each bar indicate explicitly the height of the bar. On panel \textbf{(a)} we consider the whole test set, on panel \textbf{(b)} we only consider the examples which were correctly classified by the \textit{arithmetic} ensemble average, and on panel \textbf{(c)} those correctly classified through \textit{geometric} ensemble average.
    }
    \label{fig:www}
\end{figure*}

Since the only thing that average ensembling takes into account is the confidence vectors of the models, we identify two possible reasons why EDeiTs perform better, despite the single-model performance being equal to CNNs:
\begin{itemize}
    \item[(a)] Different CNN models tend to agree on the same wrong answer more often than DeiTs.
    \item[(b)] The confidence profile of the DeiT predictions is better suited for average ensembling than the other models.
\end{itemize}
We will see that both (a) and (b) are true, though the dominant contribution comes from (a). 
In Fig.~\ref{fig:www}a we show a histogram of how many models gave a right (R) or wrong (W) prediction (\textit{e.g.} RRR denotes three correct predictions within the individual models, RRW denotes one mistake, and so on).


On Fig.~\ref{fig:www} b (and c), we show the same quantity, but restricted to the examples that were correctly classified by the arithmetic and geometric averaged ensemble models.
The CNN ensemble has more RRR cases (2523) than the  EDeiT (2430), but when the three models have some disagreement, the EDeiTs catch up and outperform the CNN ensembles.

In particular:
\begin{itemize}
    \item The correct RWW cases are 2.3x to 2.6x more common in the geometric average and arithmetic average EDeiT respectively (Geometric CNN: 10, Geometric EDeiT: 23; Arithmetic CNN: 8, Arithmetic EDeiT: 21). 
    In App.~\ref{app:RW1W2} we show that the probability that a RWW ensembling results in a correct prediction depends on the ratio between the second and third component of the ensembled confidence vector, and that the better performance of DeiT ensembles in this situation is justified by the shape of the confidence vector.
    \item The correct RRW cases are 2.4x more common in the geometric average and arithmetic average EDeiT (Geometric CNN: 93, Geometric EDeiT: 226; Arithmetic CNN: 94, Arithmetic DeiT: 225), and they represent the bulk of the improvement of DeiT versus CNN ensembles. Since the single-model performances are similar, this suggests that a higher degree of disagreement among individual models is allowing for a better ensembling.
\end{itemize}
We thus measure the mutual agreement between different models. To do so, we take the confidence vectors, $\vec c_0$, $\vec c_1$ and $\vec c_2$ of the three models, and calculate the similarity
\begin{equation}\label{eq:similarity}
    S = \frac{1}{3}(\vec c_0\cdot\vec c_1+\vec c_0\cdot\vec c_2+\vec c_1\cdot\vec c_2)\,,
\end{equation}
averaged over the full test set.
For DeiTs, we have $S=0.773\pm0.004$, while for CNNs the similarity is much higher, $S=0.945\pm0.003$. 
This is independent of which CNN models we use. If we ensemble Eff2, Eff5 and Eff6, we obtain $S=0.948\pm0.003$. 
Note that the lower correlation between predictions from different DeiT learners is even more striking given that we are comparing the \textit{same} DeiT model trained three times, with \textit{different} CNN architectures.
This suggests that the CNN predictions focus on similar sets of characteristics of the image, so when they fail, all models fail similarly. On the contrary, the predictions of separate DeiTs are more independent. Given a fixed budget of single-model correct answers, this has a double benefit: (i) the best ensembled performance is obtained by maximizing the number of RRW combinations with respect to the RRR combinations; (ii) RWW combinations result more likely in a correct answer when the two wrong answers are different (see App.~\ref{app:RW1W2}). 
The situation is analogous for geometric averaging (Fig.~\ref{fig:www}c), where we further note that there can be WWW models resulting in a correct prediction, because all the (wrong) top answers of each model can be vetoed by another model.

\paragraph*{Comparison to vanilla ViTs :}
For completeness, in App.~\ref{app:deit-vs-vit} we also provide a comparison between DeiTs~\cite{touvron:20} and vanilla ViTs~\cite{dosovitskiy:20}. Also here, we find analogous results: despite the single-model performance being similar, DeiTs ensemble better, and this can be again attributed to the lower similarity between predictions coming from independent models. This suggests that the better performance of DeiT ensembles is not related to the attention mechanism of ViTs, but rather of the distillation process which is characteristic of DeiTs (Sec.~\ref{sec:models}).


\subsection{Arithmetic versus geometric averaging for ensembling}
While arithmetic averages (AA) are most commonly used to combine predictions of classifier ensembles~\cite{alexandre:01} and yield comparably good results (see \textit{e.g.}, Refs.~\cite{alkoot:99,kittler:98}), depending on the situation geometric averaging (GA) may perform better (e.g., \cite{tax:97,mi:11}). From Fig.~\ref{fig:comparison}, it seems that GA is systematically better than AA. While this is true for EDeiTs (and ViTs), this seems not to be the case ensembling with CNN architectures.

For the \ourdata dataset, we studied AA versus GA also for the CNN and ViT architectures. As we show in Tab.~\ref{tab:Geo_and_arth_mean}, GA is better than AA in ViTs and DeiTs (especially in the F1-score), and worse than AA in CNNs. 
Since we can expect that the \textit{veto} mechanism (Sec.~\ref{sec:ensembling}) works better with less peaked confidence vectors (because with a very peaked confidence vector almost all the components are vetoed), the differences between performances of AAs and GAs for DeiTs, ViTs, and CNNs may be linked to the distribution of confidence values of the predictions. 
Indeed, both CNNs and ViTs assign considerably higher probabilities to the top-1 class than DeiTs (see Apps.~\ref{app:RW1W2},\ref{app:deit-vs-vit}), which also implies that all classes except the top-1 of each classifier suffer from a strong inhibition.

\begin{table*}[t]
\centering
\caption{Comparison of the geometric average with the arithmetic average ensembles. Details on the used models can be found in Sec.~\ref{sec:models}.
Here, DeiTs are ensembles of three DeiT-Base architectures~\cite{touvron:20}, ViT using the VIT-B16, VIT-B32, and VIT-L32 architectures~\cite{dosovitskiy:20}, and CNN using the DenseNet~\cite{huang:18}, MobileNet~\cite{sandler:18}, EfficientNet-B2~\cite{tan:19}, EfficientNet-B5~\cite{tan:19}, EfficientNet-B6~\cite{tan:19}, and EfficientNet-B7~\cite{tan:19} architectures.
}
\label{tab:Geo_and_arth_mean}
\begin{tabular}{|l|c|rr|rr|rr|}
\hline
\multirow{2}{*}{\begin{tabular}[c]{@{}l@{}}Model type\end{tabular}} & \multicolumn{1}{l|}{\multirow{2}{*}{Dataset}} & \multicolumn{2}{c|}{Arithmetic ensemble} & \multicolumn{2}{c|}{Geometric ensemble} & \multicolumn{2}{c|}{Change from AA to GA} \\ \cline{3-8} 
 & \multicolumn{1}{l|}{} & \multicolumn{1}{r|}{Accuracy} & F1-score & \multicolumn{1}{r|}{Accuracy} & F1-score & \multicolumn{1}{r|}{Accuracy} & F1-score \\ \hline
DeiT (3 models) & \multirow{3}{*}{\ourdata} & \multicolumn{1}{r|}{0.994} & 0.973 & \multicolumn{1}{r|}{0.996} & 0.984 & \multicolumn{1}{r|}{0.201\%} & 1.131\% \\ \cline{1-1} \cline{3-8} 
ViT (3 models) &  & \multicolumn{1}{r|}{0.972} & 0.922 & \multicolumn{1}{r|}{0.974} & 0.931 & \multicolumn{1}{r|}{0.206\%} & 0.976\% \\ \cline{1-1} \cline{3-8} 
CNN (6 models) &  & \multicolumn{1}{r|}{0.978} & 0.924 & \multicolumn{1}{r|}{0.977} & 0.923 & \multicolumn{1}{r|}{-0.102\%} & -0.108\% \\ \hline
\end{tabular}
\end{table*}

\section*{Discussion}
\label{sec:discussion}
We presented Ensembles of Data Efficient Image Transformers (EDeiTs) as a standard go-to method for image classification. Though the method we presented is valid for any kind of images, we provided a proof of concept of its validity with biodiversity images. Besides being of simple training and deployment (we performed no specific tuning for any of the datasets), EDeiTs systematically lead to a substantial improvement in classifying biodiversity images across all tested datasets, when compared to the previous state of the art. 
Furthermore, our results were obtained by averaging over three DeiT models, but increasing the number of individual learners can lead to a further improvement in the performances.

Focusing on a single dataset, we compared DeiT with CNN models (analogous results stem from a comparison with vanilla ViTs). Despite the similar performance of individual CNN and DeiT models, ensembling benefits DeiTs to a larger extent. We attributed this to two mechanisms. To a minor extent, the confidence vectors of DeiTs are less peaked on the highest value, which has a slight benefit on ensembling. To a major extent, independently of the architecture, the predictions of CNN models are very similar to each other (independently of whether the prediction is wrong or right), whereas different DeiTs have a lower degree of mutual agreement, which turns out beneficial towards ensembling.
This greater independence between DeiT learners also suggests that the loss landscape of DeiTs is qualitatively different from that of CNNs, and that DeiTs might be particularly suitable for algorithms that average the model weights throughout learning, such as stochastic weighted averaging~\cite{izmailov:18}, since different weight configurations seem to interpret the image in a different way. 

\begin{figure*}[tb!]
\centering
\includegraphics[width=\textwidth]{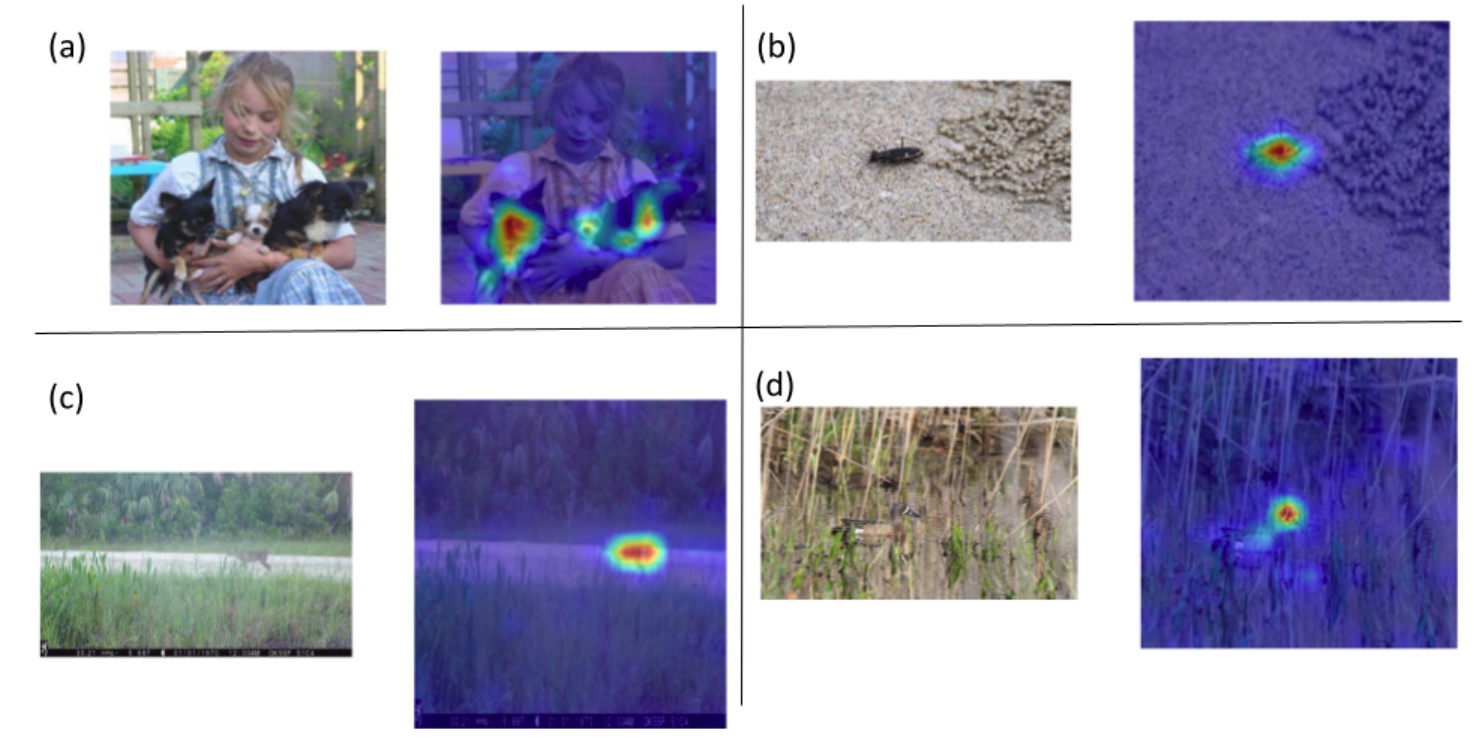}
\caption{Examples of DeiTs identifying images from different datasets:
(a) Stanford Dogs, (b) SriLankan tiger beetles, (c) Florida wild-trap, and (d) NA-Birds datasets are visualized. The original image is shown on the left in each panel, while the right reveals where our model is paying attention while classifying the species in the image.}
\label{fig:visualization}
\end{figure*}

Unlike many kinds of ViTs, the DeiT models we used have a similar number of parameters compared to CNNs, and the computational power required to train them is similar. In addition to their deployment requiring similar efforts, with higher performances, DeiTs have the additional advantage of being more straightforwardly interpretable than CNNs by ecologists, because of the attention map that characterizes transformers. The attention mechanism allows to effortlessly identify where in the image the model focused its attention (Fig.~\ref{fig:visualization}), rendering DeiTs more transparent and controllable by end users.

All these observations pose EDeiTs as a solid go-to method for the classification of ecology monitoring images. Though EDeiTs are likely to be an equally solid method also in different domains, we do not expect EDeiTs to beat the state of the art in mainstream datasets such as CIFAR~\cite{cifar} or ImageNet~\cite{deng:09}.
In fact, for such datasets, immense efforts were made to achieve the state of the art, the top architectures are heavily tailored to these datasets~\cite{recht:19}, and their training required huge numerical efforts. Even reusing those same  top architectures, it is hard to achieve high single-model performances with simple training protocols and moderate computational resources. 
In addition, while ensembling provides benefits~\cite{dascoli:20}, well-tailored architectural choices can provide the same benefits~\cite{nakkiran:20}. Therefore, it is expected that the SOTA models trained on these datasets will benefit less from ensembling.

Finally, we note that the nominal test performance of machine learning models is often subject to a decrease when the models are deployed on real world data. This phenomenon, called \textit{data shift}, can be essentially attributed to the fact that the data sets often do not adequately represent the distribution of images that is sampled at the moment of deployment~\cite{morenotorres:12}. This can be due to various reasons (sampling method, instrument degradation, seasonal effects, an so on) and is hard to harness. However, it was recently shown that Vision Transformer models (here, ViT and DeiT) are more robust to data shift~\cite{minderer:21,naseer:21,paul:22} and to other kinds of perturbations such as occlusions~\cite{naseer:21}, which is a further reason for the deployment of EDeiTs in ecological monitoring.

\section*{Methods}
\subsection{Data} \label{sec:data}
\begin{figure*}[t]
    \centering
    \includegraphics[width=.99\textwidth]{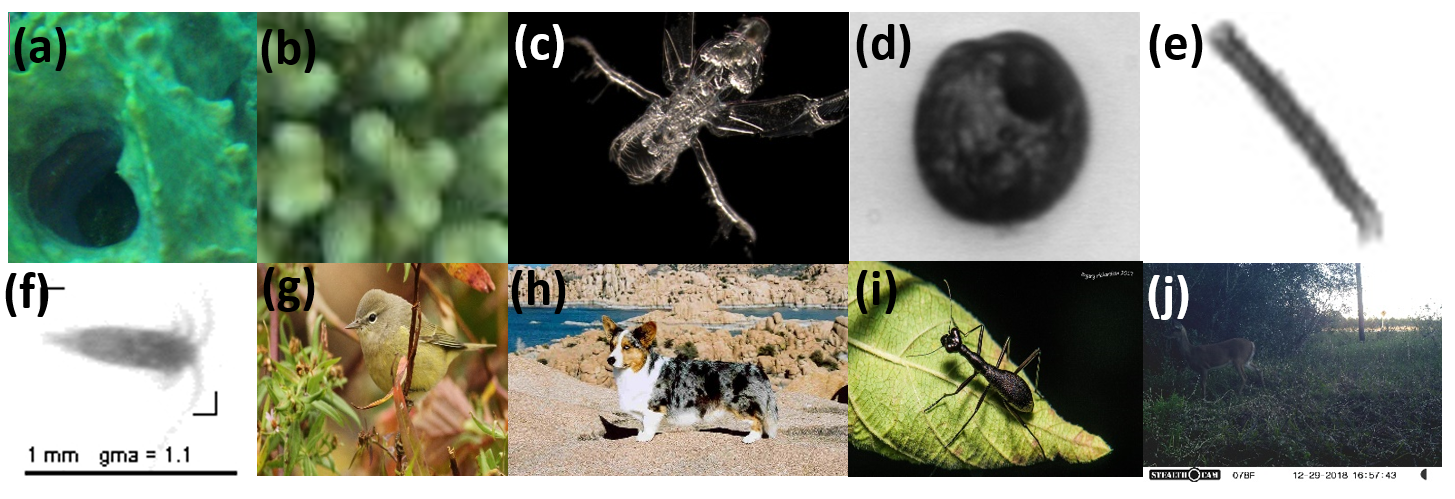}
    \caption{Examples of images from each of the datasets.(a) RSMAS (b) EILAT (c) ZooLake (d) WHOI (e) Kaggle (f) ZooScan (g) NA-Birds (h) Stanford dogs (i) SriLankan Beetles (j) Florida Wildtrap.}
    \label{fig:examples}
\end{figure*}

We tested our models on ten publicly available datasets. 
In Fig.~\ref{fig:examples} we show examples of images from each of the datasets.
When applicable, the training and test splits were kept the same as in the original dataset. For example, the ZooScan, Kaggle, EILAT, and RSMAS datasets lack a specific training and test set; in these cases, benchmarks come from $k$-fold cross-validation~\cite{zheng:17,lumini:20}, and we followed the exact same procedures in order to allow for a fair comparison.

\paragraph*{RSMAS}
This is a small coral dataset of 766 RGB image patches with a size of $256\times256$ pixels each~\cite{gomez:19}. The patches were cropped out of bigger images obtained by the University of Miami's Rosenstiel School of Marine and Atmospheric Sciences. These images were captured using various cameras in various locations. The data is separated into 14 unbalanced groups and whose labels correspond to the names of the coral species in Latin. The current SOTA for the classification of this dataset is by \cite{lumini:20}. They use the ensemble of best performing 11 CNN models. The best models were chosen based on sequential forward feature selection (SFFS) approach. Since an independent test is not available, they make use of 5-fold cross-validation for benchmarking the performances.
    
\paragraph*{EILAT} 
This is a coral dataset of 1123 64-pixel RGB image patches~\cite{gomez:19} that were created from larger images that were taken from coral reefs near Eilat in the Red sea. The image dataset is partitioned into eight classes, with an unequal distribution of data. The names of the classes correspond to the shorter version of the scientific names of the coral species. The current SOTA~\cite{lumini:20} for the classification of this dataset uses the ensemble of best performing 11 CNN models similar to RSMAS dataset and 5-fold cross-validation for benchmarking the performances. 

\paragraph*{ZooLake}
This dataset consists of 17943 images of lake plankton from 35 classes, acquired using a Dual-magniﬁcation Scripps Plankton Camera (DSPC) in Lake Greifensee (Switzerland) between 2018 and 2020~\cite{merz:21,kyathanahally:21b}. The images are colored, with a black background and an uneven class distribution. The current SOTA~\cite{kyathanahally:21} on this dataset is based on a stacking ensemble of 6 CNN models on an independent test set.

\paragraph*{WHOI}
This dataset~\cite{sosik:07} contains images of marine plankton acquired by Image FlowCytobot~\cite{olson:07}, from Woods Hole Harbor water. The sampling was done between late fall and early spring in 2004 and 2005. It contains 6600 greyscale images of different sizes, from 22 manually categorized plankton classes with an equal number of samples for each class. The majority of the classes belonging to phytoplankton at genus level. This dataset was later extended to include 3.4M images and 103 classes. The WHOI subset that we use was previously used for benchmarking plankton classification models~\cite{zheng:17,lumini:20}. The current SOTA~\cite{kyathanahally:21} on this dataset is based on average ensemble of 6 CNN models on an independent test set.

\paragraph*{Kaggle-plankton}
The original Kaggle-plankton dataset consists of plankton images that were acquired by In-situ Ichthyoplankton Imaging System (ISIIS) technology from May to June 2014 in the Straits of Florida. The dataset was published on Kaggle (https://www.kaggle.com/c/datasciencebowl) with images originating from the Hatfield Marine Science Center at Oregon State University. A subset of the original Kaggle-plankton dataset was published by~\cite{zheng:17} to benchmark the plankton classification tasks. This subset comprises of 14,374 greyscale images from 38 classes, and the distribution among classes is not uniform, but each class has at least 100 samples. The current SOTA~\cite{kyathanahally:21} uses average ensemble of 6 CNN models and benchmarks the performance using 5-fold cross-validation.

\paragraph*{ZooScan}
The ZooScan dataset consists of 3771 greyscale plankton images acquired using the Zooscan technology from the Bay of Villefranche-sur-mer~\cite{gorsky:10}. This dataset was used for benchmarking the classification models in previous plankton recognition papers~\cite{lumini:20,zheng:17}. The dataset consists of 20 classes with a variable number of samples for each class ranging from 28 to 427. The current SOTA~\cite{kyathanahally:21} uses average ensemble of 6 CNN models and benchmarks the performance using 2-fold cross-validation.

\paragraph*{NA-Birds}
NA-Birds~\cite{horn:15} is a collection of 48,000 captioned pictures of North America's 400 most often seen bird species. For each species, there are over 100 images accessible, with distinct annotations for males, females, and juveniles, totaling 555 visual categories. The current SOTA~\cite{he:21} called TransFG modifies the pure ViT model by adding contrastive feature learning and part selection module that replaces the original input sequence to the transformer layer with tokens corresponding to informative regions such that the distance of representations between confusing subcategories can be enlarged. They make use of an independent test set for benchmarking the model performances.

\paragraph*{Stanford Dogs}
The Stanford Dogs dataset comprises 20,580 color images of 120 different dog breeds from all around the globe, separated into 12,000 training images and 8,580 testing images~\cite{khosla:11}. The current SOTA~\cite{he:21} makes use of modified ViT model called TransFG as explained above in NA-Birds dataset. They make use of an independent test set for benchmarking the model performances.

\paragraph*{Sri Lankan Beetles}
The arboreal tiger beetle data~\cite{abey:21} consists of 380 images that were taken between August 2017 and September 2020 from 22 places in Sri Lanka, including all climatic zones and provinces, as well as 14 districts. \textit{Tricondyla} (3 species), \textit{Derocrania} (5 species), and \textit{Neocollyris}
 (1 species) were among the nine species discovered, with six of them being endemic . The current SOTA~\cite{abey:21} makes use of CNN-based SqueezeNet architecture and was trained using pre-trained weights of ImageNet. The benchmarking of the model performances was done on an independent test set.

\paragraph*{Florida Wild Traps}
The wildlife camera trap~\cite{gagne:21} classification dataset comprises 104,495 images with visually similar species, varied lighting conditions, skewed class distribution, and samples of endangered species, such as Florida panthers. These were collected from two locations in Southwestern Florida. These images are categorized in to 22 classes. The current SOTA~\cite{gagne:21} makes use of CNN-based ResNet-50 architecture and the performance of the model was benchmarked on an independent test set. 

\subsection{Models}\label{sec:models}
Vision transformers (ViTs)~\cite{dosovitskiy:20} are an adaptation to computer vision of the Transformers, which were originally developed for natural language processing~\cite{vaswani:17}. Their distinguishing feature is that, instead of exploiting translational symmetry, as CNNs do, they have an \textit{attention mechanism} which identifies the most relevant part of an image.
ViTs have recently outperformed CNNs in image classification tasks where vast amounts of training data and processing resources are available~\cite{vaswani:17,yufei:21}. 
However, for the vast majority of use cases and consumers, where data and/or computational resources are limiting, ViTs are essentially untrainable, even when the network architecture is defined and no architectural optimization is required.
To settle this issue, Data-efficient Image Transformers (DeiTs) were proposed~\cite{touvron:20}. These are transformer models that are designed to be trained with much less data and with far less computing resources~\cite{touvron:20}. In DeiTs, the transformer architecture has been modified to allow native distillation~\cite{zeyuan:20}, in which a student neural network learns from the results of a teacher model. Here, a CNN is used as the teacher model, 
and the pure vision transformer is used as the student network. All the DeiT models we report on here are DeiT-Base models~\cite{touvron:20}. The ViTs are ViT-B16, ViT-B32, and ViT-L32 models~\cite{dosovitskiy:20}.

\subsection{Implementation}\label{sec:implementation}
To train our models, we used transfer learning~\cite{tan:18}: we took a model that was already pre-trained on the ImageNet~\cite{deng:09} dataset, changed the last layers depending on the number of classes, and then fine-tuned the whole network with a very low learning rate.
All the models were trained with two Nvidia GTX 2080Ti GPUs.

\paragraph*{DeiTs}
We used DeiT-Base~\cite{touvron:20} architecture, using the Python package TIMM~\cite{timm}, which includes many of the well-known deep learning architectures, along with their pre-trained weights computed from the ImageNet dataset~\cite{deng:09}.
We resized the input images to 224 x 224 pixels and then, to prevent the model from overfitting at the pixel level and help it generalize better, we employed typical image augmentations during training such as horizontal and vertical flips, rotations up to 180 degrees, small zoom up's to 20\%, a small Gaussian blur, and shearing up to 10\%.
To handle class imbalance, we used class reweighting, which reweights errors on each example by how present that class is in the dataset~\cite{johnson:19}. We used sklearn utilities~\cite{pedregosa:11} to calculate the class weights which we employed during the training phase. 
    
The training phase started with a default pytorch~\cite{paszke:17} initial conditions (Kaiming uniform initializer), an AdamW optimizer with cosine annealing~\cite{loshchilov:17}, with a base learning rate of $10^{-4}$, and a weight decay value of 0.03, batch size of 32 and was supervised using cross-entropy loss. 
We trained with early stopping, interrupting training if the validation F1-score did not improve for 5 epochs. The learning rate was then dropped by a factor of 10. We iterated until the learning rate reached its final value of $10^{-6}$. This procedure amounted to around 100 epochs in total, independent of the dataset. The training time varied depending on the size of the datasets. It ranged between 20min (SriLankan Beetles) to 9h (Florida Wildtrap). We used the same procedure for all the datasets: no extra time was needed for hyperparameter tuning.

\paragraph*{ViTs}
We implemented the ViT-B16, ViT-B32 and ViT-L32 models using the Python package vit-keras (\url{https://github.com/faustomorales/vit-keras}), which includes pre-trained weights computed from the ImageNet~\cite{deng:09} dataset and the Tensorflow library~\cite{abadi:16}.

First, we resized input images to 128 x 128 and employed typical image augmentations during training such as horizontal and vertical flips, rotations up to 180 degrees, small zooms up to 20\%, small Gaussian blur, and shearing up to 10\%. To handle class imbalance, we calculated the class weights and use them during the training phase. 

Using transfer learning, we imported the pre-trained model and froze all of the layers to train the model. We removed the last layer, and in its place we added a dense layer with $n_c$ outputs (being $n_c$ the number of classes), was preceded and followed by a dropout layer. We used the Keras-tuner~\cite{omalley:19} with Bayesian optimization search~\cite{mockus:12} to determine the best set of hyperparameters, which included the dropout rate, learning-rate, and dense layer parameters (10 trials and 100 epochs). After that, the model with the best hyperparameters was trained with a default tensorflow~\cite{abadi:16} initial condition (Glorot uniform initializer) for 150 epochs using early stopping, which involved halting the training if the validation loss did not decrease after 50 epochs and retaining the model parameters that had the lowest validation loss. 

\paragraph*{CNNs}
CNNs included DenseNet~\cite{huang:18}, MobileNet~\cite{sandler:18}, EfficientNet-B2~\cite{tan:19}, EfficientNet-B5~\cite{tan:19}, EfficientNet-B6~\cite{tan:19}, and EfficientNet-B7~\cite{tan:19} architectures. We followed the training procedure described in Ref.~\cite{kyathanahally:21}, and carried out the training in tensorflow. 


\subsection{Ensemble learning}\label{sec:ensembling}
We adopted average ensembling, which takes the confidence vectors of different learners, and produces a prediction based on the average among the confidence vectors. With this procedure, all the individual models contribute equally to the ﬁnal prediction, irrespective of their validation performance.
Ensembling usually results in superior overall classification metrics and model robustness~\cite{seni:10,zhang:12}.

Given a set of $n$ models, with prediction vectors $\vec c_i~(i=1,\ldots,n)$, these are typically aggregated through an arithmetic average. The components of the ensembled confidence vector $\vec c_{AA}$, related to each class $\alpha$ are then
\begin{equation}\label{eq:arithmetic}
 c_{AA,\alpha} = \frac{1}{n}\sum_{i=1}^n c_{i,\alpha}\,.
\end{equation}
Another option is to use a geometric average,
\begin{equation}\label{eq:geometric}
c_{GA,\alpha} = \sqrt[n]{\prod_{i=1}^n c_{i,\alpha}}\,.
\end{equation}
We can normalize the vector $\vec c_g$, but this is not relevant, since we are interested in its largest component, $\displaystyle\max_\alpha(c_{GA,\alpha})$, and normalization affects all the components in the same way. As a matter of fact, also the $n^\mathrm{th}$ root does not change the relative magnitude of the components, so instead of $\vec c_{GA}$ we can use a product rule: $\displaystyle\max_\alpha(c_{GA,\alpha})=\max_\alpha(c_{PROD,\alpha})$, with $\displaystyle c_{PROD,\alpha} = \prod_{i=1}^n c_{i,\alpha}$.




While these two kinds of averaging are equivalent in the case of two models and two classes, they are generally different in any other case~\cite{alexandre:01}.
For example, it can easily be seen that the geometric average penalizes more strongly the classes for which at least one learner has a very low confidence value, a property that was termed \textit{veto} mechanism~\cite{tax:97} (note that, while in Ref.~\cite{tax:97} the term \textit{veto} is used when the confidence value is exactly zero, here we use this term in a slightly looser way).

\section*{Acknowledgements}
This project was funded by the Eawag DF project Big-Data Workflow (\#5221.00492.999.01), the Swiss Federal Office for the Environment (contract Nr Q392-1149) and the Swiss National Science Foundation (project 182124).

\section*{Author contributions statement}
M.B.J. designed the study,
S.K. mined the data, S.K. built the models, S.K., T.H., E.M., T.B., M.R., P.B., F.P. and M.B.J. were actively involved in the discussion while building and improving the models and data, S.K. and M.B.J. wrote the paper. All the authors contributed to the manuscript.

\section*{Competing interests}
The authors declare that the research was conducted in the absence of any commercial or financial relationships that could be construed as a potential conflict of interest.

\section*{Code Availability Statement}
The code for the reproduction of our results is available at 
\url{github.com/kspruthviraj/Plankiformer}.

\section*{Data Availability Statement}
All the data we used is open access.
The datasets analysed during the current study are available in the repositories, that we indicate in Sec.~\ref{sec:data}.

\appendix

\section{Comparisons of DeiT ensembles with previous SOTA}\label{app:comparisons-sota}
Here, we provide further comparisons between EDeiTs and the previous SOTA.
In Tab.~\ref{tab:performances} and Tab.~\ref{tab:performances_geo} we show both the model accuracy and the (macro) F1-score, in order to have both micro- and macro-averaged descriptors (\textit{i.e.} descriptors which are directly influenced by data imbalance and that are not). Both improve systematically with respect to the SOTA. In particular, the improvement in the F1 scores indicates that the classification of the minority classes improved.
Since the amount of improvement from the SOTA is bounded by the error of the SOTA, we find that a better metric to define the improvement of our models is the error, defined as the fraction of misclassified examples (\textit{i.e.} 1-accuracy). As shown in the rightmost part of Tab.~\ref{tab:performances}, for average ensembling the error is reduced widely, from a 18.48\% decrease in the NA-Birds dataset (where the SOTA models are a kind of ViTs, TransFG~\cite{he:21}), to as much of an 87.5\% error reduction in the RSMAS dataset, depicting corals (where the SOTA models are ensembles of convolutional networks).
The performance increase in both accuracy and F1 score is even sharper when using geometric averaging (Tab.~\ref{tab:performances_geo}), where the error reduction  ranges from 29.35\% to 100\%. A 100\% error reduction means that none of the test images were misclassified by the EDeiT.
\begin{table*}[t]
\caption{Summary of the performances of EDeiTs (combining learners through an arithmetic mean) and the current SOTA models on public datasets.}
\label{tab:performances}
\resizebox{\textwidth}{!}{\begin{tabular}{|l|l|l|l|lll|lll|lll|}
\hline
\multirow{2}{*}{Dataset name} & \multirow{2}{*}{Image types} & \multirow{2}{*}{Classes} & \multirow{2}{*}{\begin{tabular}[c]{@{}l@{}}No. of \\ images\end{tabular}} & \multicolumn{3}{c|}{\begin{tabular}[c]{@{}c@{}}Previous State of the Art\end{tabular}} & \multicolumn{3}{c|}{\begin{tabular}[c]{@{}c@{}}Arithmetic average EDeiTs\end{tabular}} & \multicolumn{2}{c|}{\begin{tabular}[c]{@{}c@{}}Absolute improvement\end{tabular}} &\multicolumn{1}{c|}{\begin{tabular}[c]{@{}c@{}}Relative \\ improvement\end{tabular}} \\ \cline{5-13} 
 &  &  &  & \multicolumn{1}{l|}{Accuracy} & \multicolumn{1}{l|}{F1 Score} & Error & \multicolumn{1}{l|}{Accuracy} & \multicolumn{1}{l|}{F1-Score} & Error & \multicolumn{1}{l|}{Accuracy} & \multicolumn{1}{l|}{F1-Score} & Error \\ \hline
RSMAS \cite{gomez:19} & Coral-reef & 14 & 766 & \multicolumn{1}{l|}{0.992 \cite{lumini:20}} & \multicolumn{1}{l|}{0.995} & 0.008 & \multicolumn{1}{l|}{0.999} & \multicolumn{1}{l|}{0.999} & 0.001 & \multicolumn{1}{l|}{0.70\%} & \multicolumn{1}{l|}{0.40\%} & -87.50\% \\ \hline
EILAT \cite{gomez:19} & Coral-reef & 8 & 1123 & \multicolumn{1}{l|}{0.989 \cite{lumini:20}} & \multicolumn{1}{l|}{0.990} & 0.011 & \multicolumn{1}{l|}{0.997} & \multicolumn{1}{l|}{0.997} & 0.003 & \multicolumn{1}{l|}{0.80\%} & \multicolumn{1}{l|}{0.70\%} & -72.73\% \\ \hline
ZooLake \cite{kyathanahally:21b} & Plankton & 35 & 17943 & \multicolumn{1}{l|}{0.979 \cite{kyathanahally:21}} & \multicolumn{1}{l|}{0.930} & 0.021 & \multicolumn{1}{l|}{0.994} & \multicolumn{1}{l|}{0.973} & 0.006 & \multicolumn{1}{l|}{1.50\%} & \multicolumn{1}{l|}{4.60\%} & -71.43\% \\ \hline
WHOI ~\cite{sosik:07} & Plankton & 22 & 6600 & \multicolumn{1}{l|}{0.961 \cite{kyathanahally:21}} & \multicolumn{1}{l|}{0.961} & 0.039 & \multicolumn{1}{l|}{0.995} & \multicolumn{1}{l|}{0.995} & 0.005 & \multicolumn{1}{l|}{3.40\%} & \multicolumn{1}{l|}{3.40\%} & -87.18\% \\ \hline
Kaggle ~\cite{zheng:17} & Plankton & 38 & 14374 & \multicolumn{1}{l|}{0.947 \cite{kyathanahally:21}} & \multicolumn{1}{l|}{0.937} & 0.053 & \multicolumn{1}{l|}{0.986} & \multicolumn{1}{l|}{0.985} & 0.014 & \multicolumn{1}{l|}{3.90\%} & \multicolumn{1}{l|}{4.80\%} & -73.58\% \\ \hline
ZooScan ~\cite{gorsky:10} & Plankton & 20 & 3771 & \multicolumn{1}{l|}{0.898 \cite{kyathanahally:21}} & \multicolumn{1}{l|}{0.915} & 0.102 & \multicolumn{1}{l|}{0.965} & \multicolumn{1}{l|}{0.975} & 0.035 & \multicolumn{1}{l|}{6.70\%} & \multicolumn{1}{l|}{6.00\%} & -65.69\% \\ \hline
NA-Birds ~\cite{horn:15} & Birds & 555 & 48562 & \multicolumn{1}{l|}{0.908 \cite{he:21}} & \multicolumn{1}{l|}{-} & 0.092 & \multicolumn{1}{l|}{0.925} & \multicolumn{1}{l|}{0.906} & 0.075 & \multicolumn{1}{l|}{1.70\%} & \multicolumn{1}{l|}{-} & -18.48\% \\ \hline
\begin{tabular}[c]{@{}l@{}}Stanford \\ Dogs ~\cite{khosla:11}\end{tabular} & Dogs & 120 & 20580 & \multicolumn{1}{l|}{0.923~\cite{he:21}} & \multicolumn{1}{l|}{-} & 0.077 & \multicolumn{1}{l|}{0.961} & \multicolumn{1}{l|}{0.958} & 0.039 & \multicolumn{1}{l|}{3.80\%} & \multicolumn{1}{l|}{-} & -49.35\% \\ \hline
\begin{tabular}[c]{@{}l@{}}Sri Lankan \\ Beetles ~\cite{abey:21}\end{tabular} & Tiger beetles & 9 & 361 & \multicolumn{1}{l|}{0.910 ~\cite{abey:21}} & \multicolumn{1}{l|}{-} & 0.090 & \multicolumn{1}{l|}{0.932} & \multicolumn{1}{l|}{0.919} & 0.068 & \multicolumn{1}{l|}{2.20\%} & \multicolumn{1}{l|}{-} & -24.44\% \\ \hline
\begin{tabular}[c]{@{}l@{}}Florida \\ Wild trap ~\cite{gagne:21}\end{tabular} & \begin{tabular}[c]{@{}l@{}}Wild animals \\ \& birds\end{tabular} & 22 & 104495 & \multicolumn{1}{l|}{0.790 ~\cite{gagne:21}} & \multicolumn{1}{l|}{-} & 0.210 & \multicolumn{1}{l|}{0.928} & \multicolumn{1}{l|}{0.613} & 0.072 & \multicolumn{1}{l|}{13.80\%} & \multicolumn{1}{l|}{-} & -65.71\% \\ \hline
\end{tabular}}
\end{table*}

\begin{table*}[t]
\caption{Summary of the performances of EDeiTs (combining learners through geometric mean) and the current SOTA models on public datasets.}
\label{tab:performances_geo}
\resizebox{\textwidth}{!}{\begin{tabular}{|l|l|l|l|lll|lll|lll|}
\hline
\multirow{2}{*}{Dataset name} & \multirow{2}{*}{Image types} & \multirow{2}{*}{Classes} & \multirow{2}{*}{\begin{tabular}[c]{@{}l@{}}No. of \\ images\end{tabular}} & \multicolumn{3}{c|}{\begin{tabular}[c]{@{}c@{}}Previous State of the Art\end{tabular}} & \multicolumn{3}{c|}{\begin{tabular}[c]{@{}c@{}}Geometric average EDeiTs\end{tabular}} & \multicolumn{2}{c|}{\begin{tabular}[c]{@{}c@{}}Absolute improvement\end{tabular}} &\multicolumn{1}{c|}{\begin{tabular}[c]{@{}c@{}}Relative \\ improvement\end{tabular}} \\ \cline{5-13} 
 &  &  &  & \multicolumn{1}{l|}{Accuracy} & \multicolumn{1}{l|}{F1 Score} & Error & \multicolumn{1}{l|}{Accuracy} & \multicolumn{1}{l|}{F1-Score} & Error & \multicolumn{1}{l|}{Accuracy} & \multicolumn{1}{l|}{F1-Score} & Error \\ \hline
RSMAS \cite{gomez:19} & Coral-reef & 14 & 766 & \multicolumn{1}{l|}{0.992 \cite{lumini:20}} & \multicolumn{1}{l|}{0.995} & 0.008 & \multicolumn{1}{l|}{1.000} & \multicolumn{1}{l|}{1.000} & 0.000 & \multicolumn{1}{l|}{0.80\%} & \multicolumn{1}{l|}{0.50\%} & -100.00\% \\ \hline
EILAT \cite{gomez:19} & Coral-reef & 8 & 1123 & \multicolumn{1}{l|}{0.989 \cite{lumini:20}} & \multicolumn{1}{l|}{0.990} & 0.011 & \multicolumn{1}{l|}{0.997} & \multicolumn{1}{l|}{0.997} & 0.003 & \multicolumn{1}{l|}{0.80\%} & \multicolumn{1}{l|}{0.70\%} & -72.73\% \\ \hline
ZooLake \cite{kyathanahally:21b} & Plankton & 35 & 17943 & \multicolumn{1}{l|}{0.979 \cite{kyathanahally:21}} & \multicolumn{1}{l|}{0.930} & 0.021 & \multicolumn{1}{l|}{0.996} & \multicolumn{1}{l|}{0.984} & 0.003 & \multicolumn{1}{l|}{1.80\%} & \multicolumn{1}{l|}{5.70\%} & -80.9\% \\ \hline
WHOI ~\cite{sosik:07} & Plankton & 22 & 6600 & \multicolumn{1}{l|}{0.961 \cite{kyathanahally:21}} & \multicolumn{1}{l|}{0.961} & 0.039 & \multicolumn{1}{l|}{0.995} & \multicolumn{1}{l|}{0.995} & 0.005 & \multicolumn{1}{l|}{3.40\%} & \multicolumn{1}{l|}{3.40\%} & -87.18\% \\ \hline
Kaggle ~\cite{zheng:17} & Plankton & 38 & 14374 & \multicolumn{1}{l|}{0.947 \cite{kyathanahally:21}} & \multicolumn{1}{l|}{0.937} & 0.053 & \multicolumn{1}{l|}{0.990} & \multicolumn{1}{l|}{0.989} & 0.010 & \multicolumn{1}{l|}{4.30\%} & \multicolumn{1}{l|}{5.20\%} & -81.13\% \\ \hline
ZooScan ~\cite{gorsky:10} & Plankton & 20 & 3771 & \multicolumn{1}{l|}{0.898 \cite{kyathanahally:21}} & \multicolumn{1}{l|}{0.915} & 0.102 & \multicolumn{1}{l|}{0.976} & \multicolumn{1}{l|}{0.984} & 0.024 & \multicolumn{1}{l|}{7.80\%} & \multicolumn{1}{l|}{6.90\%} & -76.47\% \\ \hline
NA-Birds ~\cite{horn:15} & Birds & 555 & 48562 & \multicolumn{1}{l|}{0.908 \cite{he:21}} & \multicolumn{1}{l|}{-} & 0.092 & \multicolumn{1}{l|}{0.935} & \multicolumn{1}{l|}{0.918\%} & 0.065 & \multicolumn{1}{l|}{2.70\%} & \multicolumn{1}{l|}{-} & -29.35\% \\ \hline
\begin{tabular}[c]{@{}l@{}}Stanford \\ Dogs ~\cite{khosla:11}\end{tabular} & Dogs & 120 & 20580 & \multicolumn{1}{l|}{0.923 \cite{he:21}} & \multicolumn{1}{l|}{-} & 0.077 & \multicolumn{1}{l|}{0.972} & \multicolumn{1}{l|}{0.970} & 0.028 & \multicolumn{1}{l|}{4.90\%} & \multicolumn{1}{l|}{-} & -63.64\% \\ \hline
\begin{tabular}[c]{@{}l@{}}Sri Lankan \\ Beetles ~\cite{abey:21}\end{tabular} & Tiger beetles & 9 & 361 & \multicolumn{1}{l|}{0.910 ~\cite{abey:21}} & \multicolumn{1}{l|}{-} & 0.090 & \multicolumn{1}{l|}{0.932} & \multicolumn{1}{l|}{0.912} & 0.068 & \multicolumn{1}{l|}{2.20\%} & \multicolumn{1}{l|}{-} & -24.44\% \\ \hline
\begin{tabular}[c]{@{}l@{}}Florida \\ Wild trap ~\cite{gagne:21}\end{tabular} & \begin{tabular}[c]{@{}l@{}}Wild animals \\ \& birds\end{tabular} & 22 & 104495 & \multicolumn{1}{l|}{0.790 ~\cite{gagne:21}} & \multicolumn{1}{l|}{-} & 0.210 & \multicolumn{1}{l|}{0.932} & \multicolumn{1}{l|}{0.627} & 0.068 & \multicolumn{1}{l|}{14.20\%} & \multicolumn{1}{l|}{-} & -67.62\% \\ \hline
\end{tabular}}
\end{table*}

\section{Performances on ZooLake}\label{app:zoolake}
\begin{figure}[bht]
\centering
\includegraphics[width=.99\columnwidth]{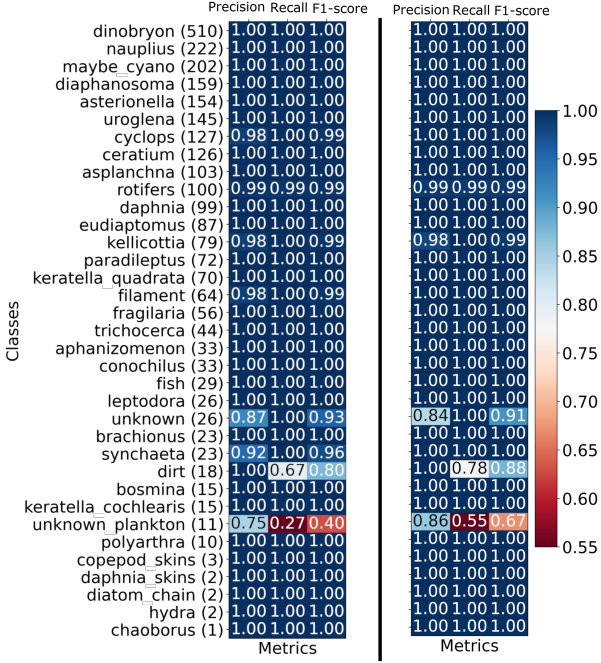}
\caption{Per-Class precision, recall and F1-score of EDeiT models on the ZooLake dataset. On the left side, we have the arithmetic average EDeiT, and on the right we have the geometric average. The number within brackets indicates the number of test images of the corresponding class.}
\label{fig:zoolake}
\end{figure}

\begin{figure*}[tbh]
\centering
\includegraphics[width=0.99\textwidth]{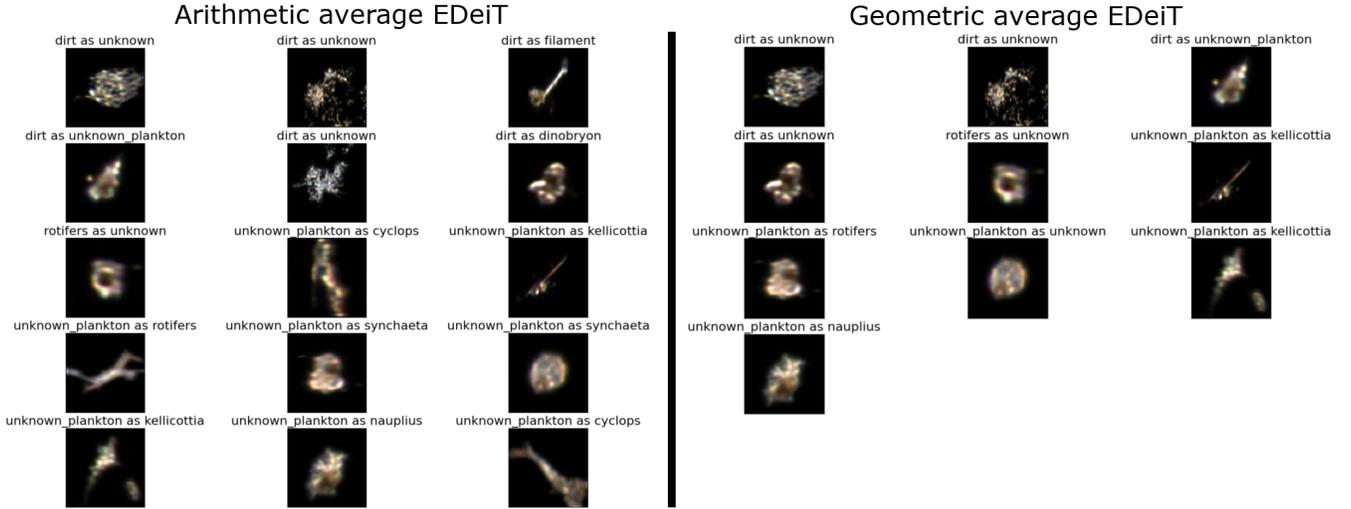}
\caption{{Misclassified images of the EDeiT models on the ZooLake dataset. On the left side, we have the misclassifications from arithmetic average EDeiT, and on the right from geometric average EDeiT. All the misclassifications involve junk classes and all the images misclassified by Geometric average EDeiT were also misclassified by arithmetic average EDeiT.}}
\label{fig:misclassification}
\end{figure*}

In Fig.~\ref{fig:zoolake} we show the per-class performances of the ensembled DeiT models on the ZooLake dataset. The model misclassified 15 out of 2691 test images. Of those, 14 belonged to junk/container categories that are not univocally defined (\textit{i.e.} there is a degree of arbitrariness when defining and labeling the junk classes). In Fig.~\ref{fig:misclassification} we explicitly show the 15 misclassified images.

\section{Arithmetic Average Ensembling in the RWW case}\label{app:RW1W2}
\begin{figure}[tbh]
\centering
\includegraphics[width=\columnwidth]{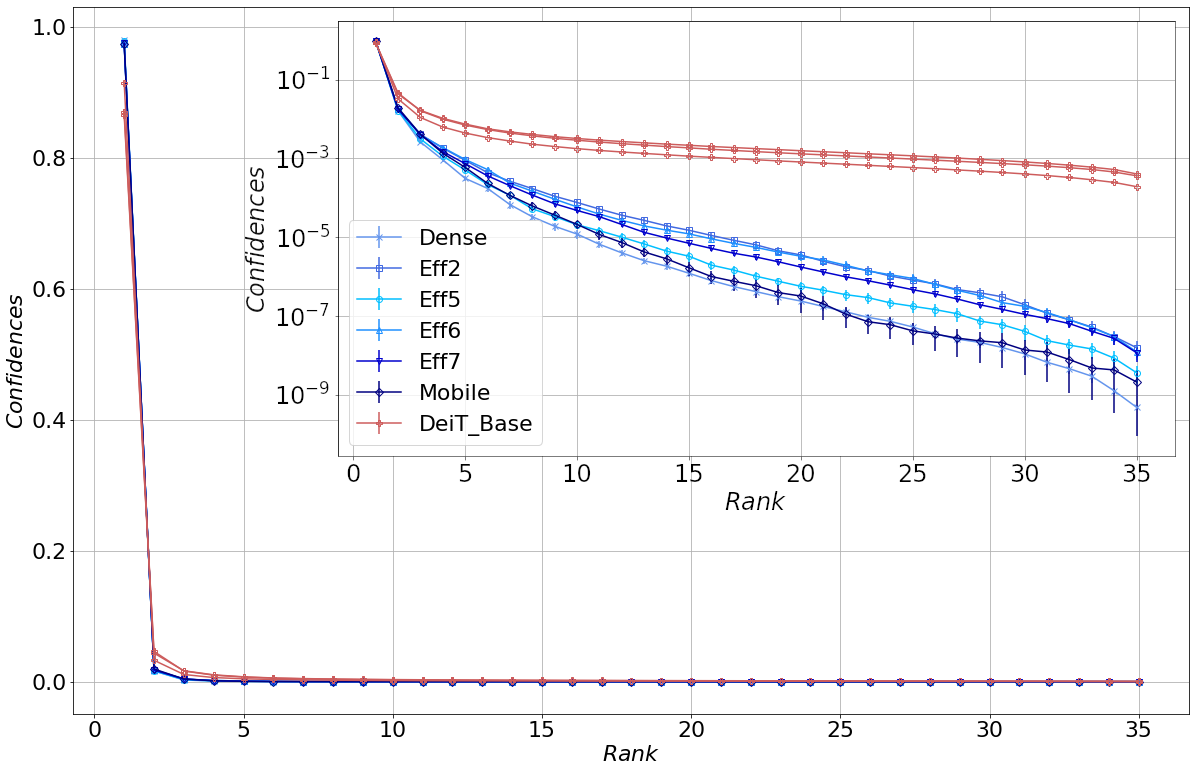}
\caption{For various model types, the mean confidences (y-axis) across all classes (x-axis) are shown. When it comes to classification, the CNN models (blues) are more confident than the DeiT models (red), as can be observed in the outset figure (linear scale). The inset figure shows the same in log scale.}
\label{fig:confs}
\end{figure}

Here, we show that also when the correct guesses are not a majority, DeiT models tend to exhibit better arithmetic average ensembling.
In Fig.~\ref{fig:confs} we show the sorted average confidence vector of the architectures we are considering. We see that all CNNs have a confidence profile which is significantly different from the DeiTs. We can thus assume that CNNs and DeiTs have two different confidence profiles, and call them $\vec C_\mathrm{CNN}$ and $\vec C_\mathrm{DeiT}$. The first component of these vectors, the largest one, $C_0$, indicates the class that is guessed by the model.

\begin{figure*}
    \centering
    \includegraphics[width=.99\textwidth]{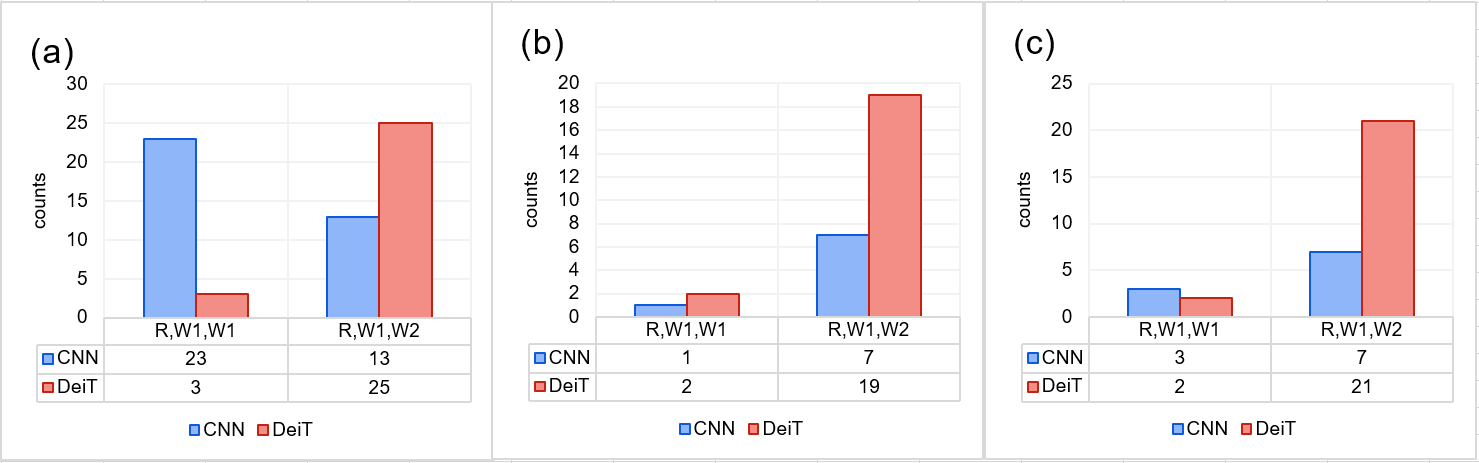}
    \caption{Comparison between ensembled CNNs (EfficientNet B7, MobileNet, DenseNet) with ensembled DeiTs. 
    \textbf{(a)}: The bars show how many times, out of the full ZooLake test set, the single learners within the ensemble model gave one correct answer (R) and two wrong answers that were indicating the same class ($W_1$ and $W_1$). The bars indicate one R answer and two wrong answers that differ from each other ($W_1$ and $W_2$).
     \textbf{(b)}: Same as (a), but only restricted to the examples that resulted in a correct classification by the arithmetic average EDeiT model. {\textbf{(c)}: Same as (a), but only restricted to the examples that resulted in a correct classification by the geometric average EDeiT model.}}
    \label{fig:mistake-cases}
\end{figure*}

Since these vectors are highly peaked on $C_0$ (so $C_0-C_1$ is large), we can state that, if an example is misclassified, then the single-model guesses will most likely be:
\begin{itemize}
    \item All wrong ($W, W, W$). Since $C_0-C_1$ is large, we cannot get a right ensemble prediction if all the models are wrong. This is true for all architectures, so this case cannot explain the better ensembling of the DeiTs.
    \item One right, and two wrong, with both wrong models guessing the same class ($R, W_1, W_1$). If both wrong guesses fall on the same class, the ensembled guess is wrong with high probability. 
    Given the higher similarity among CNN confidences [Eq.~(1) in the main text], we expect this situation to be more rare for DeiT models. We show this in Fig.~\ref{fig:mistake-cases}a, where we see that the ($R, W_1, W_1$) cases are {almost} 8 times more common in the CNNs. Comparing with Fig.~\ref{fig:mistake-cases}b, we see that most of these cases result in a mistake from the classifier. Therefore, the lower similarity between independent DeiT predictions implies better ensembling also in the $(R, W_1, W_1)$ case.
    \item One right, and two wrong, with the two wrong guesses being on different classes ($R, W_1, W_2$). As shown in Fig.~\ref{fig:mistake-cases}a, this occurs 
    slightly more often with DeiTs, which is reasonable, given that DeiTs are more likely than CNNs to provide three different predictions. As we will show, in this situation, the probability that the classifier gives the correct answer depends on the shape of the confidence vector, and that of the DeiT gives better performances.
\end{itemize}

\subsection*{How the confidence vectors influence ensembling}
We now show how the shape of the sorted confidence vectors is influences ensembling in the ($R,W_1,W_2$) case.
As shown in Fig.~\ref{fig:confs}, the confidence profiles $\vec C$ vary depending on the model class. Since the confidence vector does not have the same exact profile for every image the model sees, we assume that these are Gaussian and
define a standard deviation (not standard error) vector $\vec \sigma$, which defines how much each component of $\vec C$ fluctuates around its central value.

\begin{figure}[t]
\centering
\begin{subfigure}[b]{\columnwidth}
   \includegraphics[width=1\linewidth]{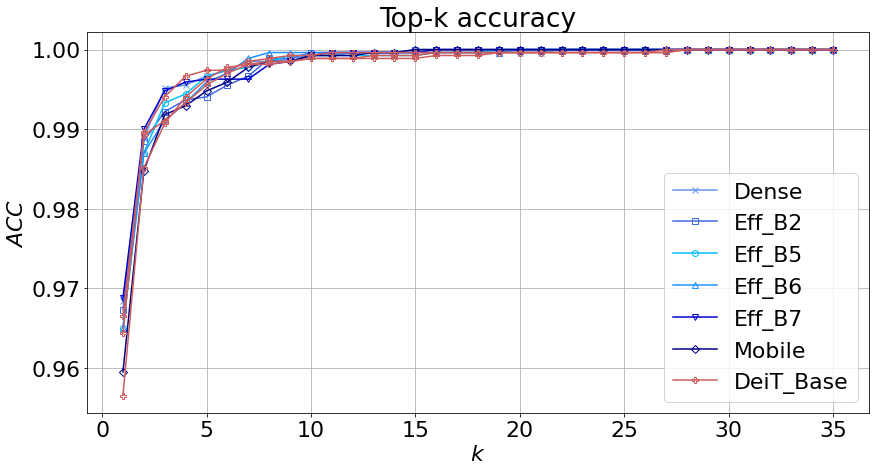}
   \caption{}
   \label{fig:topk} 
\end{subfigure}

\begin{subfigure}[b]{\columnwidth}
   \includegraphics[width=1\linewidth]{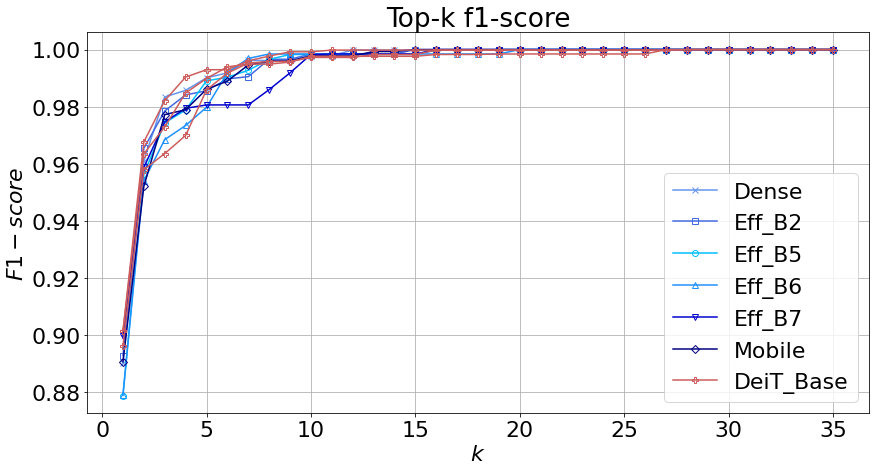}
   \caption{}
   \label{fig:topkf}
\end{subfigure}

\caption[]{(a) Top-$k$ accuracy vs $k$ and (b) Top-$k$ f1-score vs $k$ for the several models tested on the ZooLake dataset.}
\end{figure}

The components of $\C$ indicate the probability that the model assigns to the classes. These roughly correlate with the true probabilities.
We see this from Fig.~\ref{fig:topk}, which shows the top-$k$ accuracy, $A_k$, as a function of $k$. $A_k$ is the accuracy that we get if we define that a prediction is correct if any of the top $k$ predictions is correct. We see that $A_k$ starts at a value similar to $C_0$, and quickly grows, reaching 1 for small $k$.

In other words, when $C_0$ indicates the wrong class, it is most likely that $C_1$ is the correct class, and so on. 

Let us call $m_0, m_1$ and $m_2$ the three models that are used for ensembling. The confidence profiles of these three models, which we will call $\vec c^0, \vec c^1$ and $\vec c^2$, will follow the Gaussian process defined by $\C$ and $\s$, but the classes to which each of the confidences is assigned will vary according to the model. For concreteness, let us postulate that for a given image, depicting class $A$, $m_0$ provides a correct prediction. We can posit $\vec c^0=\C$, and order the classes according to how they were scored by $m_0$.
So, the confidence assigned to class $A$ is $c^0(A)=C_0$ is the confidence assigned to class $A$ by $m_0$; $c^0(B)=C_1$ is the confidence assigned to class $B$, and so on.

Let us now pass to $m_1$. Since by hypothesis it gives a wrong prediction, $c^1(A)\neq C_0$. Since when a model prediction is wrong, the second-ranked confidence is the most likely to be correct, typically $c^1(A)=C_1$. Model $m_1$ will assign $C_0$ to any of the remaining classes. In the most unfavorable cases, class $B$ is very similar to class $A$, implying $c^1(B)=C_0$; and class $C$ is also not too different, so $c^1(C)=C_2$.
An equivalent reasoning for $m_2$ leads to $c^2(A)=C_1$, $c^2(B)=C_2$ and  $c^2(C)=C_0$.

The predictions of the ensembled model are the averages across the three models,
\begin{align}
\label{eq:censA}
    c^\mathrm{(ENS)}(A) &=\frac{1}{3}(C_0+2C_1)\,,\\
\label{eq:censB}
    c^\mathrm{(ENS)}(B) &=\frac{1}{3}(C_0+C_1+C_2)\,,\\
\label{eq:censC}
    c^\mathrm{(ENS)}(C) &=\frac{1}{3}(C_0+2C_2)\,.
\end{align}
From Eqs.~\eqref{eq:censA}, \eqref{eq:censB} and~\eqref{eq:censC} we see that in case of model disagreement, ensembling tends to choose the correct answer. However, it is also possible that because of the fluctuations of the confidences around their central value, $c^\mathrm{(ENS)}(B)$ (and $c^\mathrm{(ENS)}(C)$) become larger than $c^\mathrm{(ENS)}(A)$, and the ensembled model misclassifies. The probability of a misclassification due to fluctuations is 
\begin{widetext}
\begin{align}
\label{eq:prob-misB}
    P(c^\mathrm{(ENS)}(A)<c^\mathrm{(ENS)}(B)) =& \frac12\left[1+\erf\left(\frac{C_2-C_1}{\sqrt{2(2\sigma_0^2+3\sigma_1^2+\sigma_2^2)}}\right)\right]\,,
    \\
\label{eq:prob-misC}
    P(c^\mathrm{(ENS)}(A)<c^\mathrm{(ENS)}(C)) =& 
    \frac12\left[1+\erf\left(\frac{2(C_2-C_1)}{\sqrt{2(2\sigma_0^2+2\sigma_1^2+2\sigma_2^2)}}\right)\right]\,.
\end{align}
\end{widetext}
Eqs.~\eqref{eq:prob-misB} and~\eqref{eq:prob-misC} reflect common intuition: the ensembled classifier is maximally efficient when the fluctuations are small and the difference $C_1-C_2$ is big. Thus, to leading order, the comparison of the ensembled classifiers in the ($R,W_1,W_2$) situation boils down to a comparison of the ratio
\begin{equation}
    R=\frac{(C_1-C_2)}{\sqrt{2\sigma_0^2+3\sigma_1^2+\sigma_2^2 } }
\end{equation}
related to each model. A larger $R$ implies a lower error. 
For ensembles of CNNs, this ratio is $R^\mathrm{(CNN)}=0.08$, while for DeiT it is $R^\mathrm{(DeiT)}=0.10$, which translate into $P^\mathrm{(CNN)}=0.53$ and $P^\mathrm{(DeiT)}=0.54$. 
This is a small difference, but the main point is that it goes in the same direction of the $(R,W_1,W_1)$ contribution.

\section{Comparing DeiT with ViT}\label{app:deit-vs-vit}

We compare DeiT with ViT models on the ZooLake dataset.
In Tab.~\ref{tab:individual-performance-vit} we compare the performance of ViT with DeiT models. 
When we take the single models, DeiTs do not perform better than ViTs. In particular, the single-model performance of ViT-B16 models outperforms DeiTs. However, when ensembling is carried out, DeiTs perform much better, both in accuracy and in F1-score.
\begin{table*}[t]
\centering
\caption{Summary of the performance of the individual models on the ZooLake dataset. The ensemble score on the rightmost column is obtained by averaging across 3 different initial conditions. The ViT\_3\_avg model is an ensemble of the best of each ViT-B16, ViT-B32 \& ViT-L32 models . The numbers in parentheses are the standard errors, referred to the last significant digit.}
\label{tab:individual-performance-vit}
\begin{tabular}{|l|l|l|l|l|l|}
\hline
\textbf{Type} & \textbf{\begin{tabular}[c]{@{}l@{}}No. of \\ params \\ for each \\ model\end{tabular}} & \textbf{\begin{tabular}[c]{@{}l@{}}Accuracy\\ Mean\end{tabular}} & \textbf{\begin{tabular}[c]{@{}l@{}}F1-score \\ Mean\end{tabular}} & \textbf{\begin{tabular}[c]{@{}l@{}}Arithmetic \\ Average \\Ensemble   \\ (accuracy/\\ F1-score)\end{tabular}} & \textbf{\begin{tabular}[c]{@{}l@{}}Geometric \\ Average\\ Ensemble   \\ (accuracy/\\ F1-score)\end{tabular}} \\ \hline
\textbf{ViT-B16} & 85.7M & 0.973(1) & 0.918(2) & 0.976/0.919 & 0.976/0.921\\ \hline
\textbf{ViT-B32} & 87.5M & 0.960(2) & 0.886(6) & 0.966/0.893 & 0.964/0.889\\ \hline
\textbf{ViT-L32} & 305.5M & 0.960(2) & 0.894(2) & 0.967/0.908 & 0.966/0.903\\ \hline
\textbf{ViT\_3\_avg} & - & - & - & 0.972/0.922 & 0.974/0.931\\ \hline
\textbf{DeiT-Base} & 85.8M & 0.962(3) & 0.899(2) & 0.994/0.973 & 0.996/0.984\\ \hline
\end{tabular}
\end{table*}
In addition to ensembling ViTs over initial conditions, we also take the best ViT models from each of the architectures and make a 3-model ensemble (ViT\_3\_avg) over those (Tab.~\ref{tab:individual-performance-vit}). The ensemble model with 3 different architectures is similar to performance obtained by ensembling three B-16 models. If we ensemble 5 models (three B16, one B32 and one L32) models then the F1-score slightly improves to 0.930 compared to 0.922 of ViT\_3\_avg. However, the ensemble of three DeiTs outperforms even the ensemble of 5 ViTs.

This better generalization stems from a major mutual independence of individual learners. This can be seen from the similarity between confidence vectors [Eq.~(1) in the main text] of ViTs versus DeiTs.
While for DeiTs we have $S=0.773\pm0.004$, the similarity of ViTs is much higher ($S=0.969\pm0.002$ for ViT-B16, $S=0.955\pm0.003$ for ViT-B32, $S=0.954\pm0.003$ for ViT-L32, and $S=0.956\pm0.003$ for the ensemble over different ViT architectures).

\begin{figure*}[t]
    \centering
    \includegraphics[width=.99\textwidth]{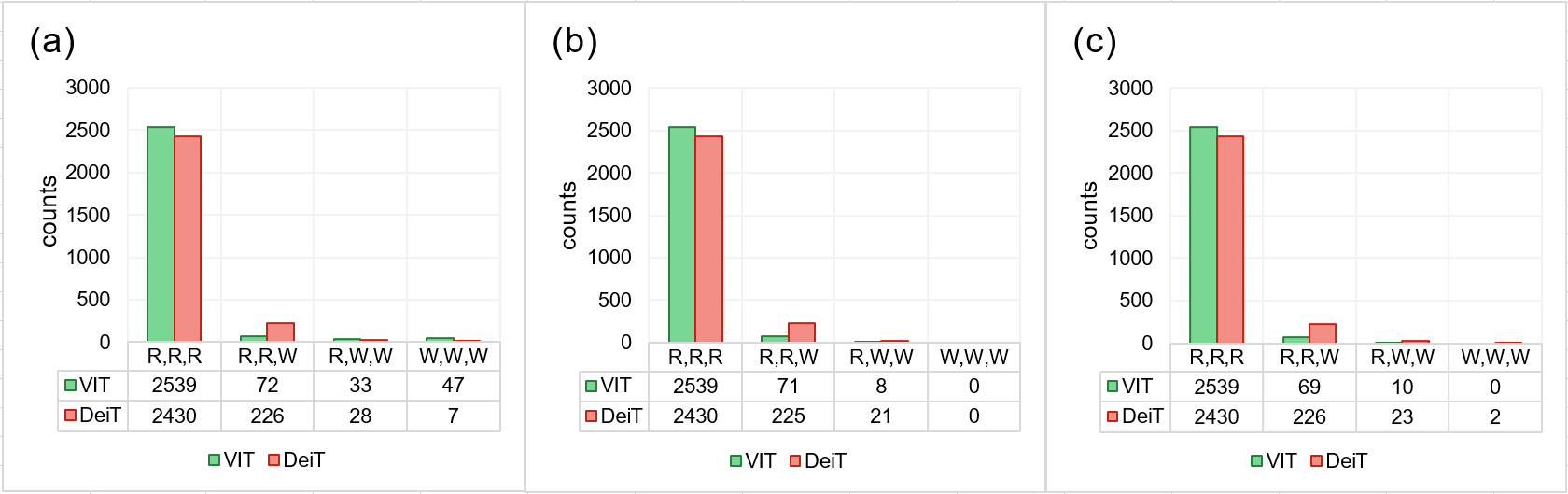}
    \caption{Comparison between 3-model ensemble models based on VITs (B16, B32 and L32) and on DeiTs on the ZooLake test set. The bar heights indicate how often each combination (RRR, RRW, RWW, WWW) appeared. RRR indicates that all the models gave the right answer, RRW means that one model gave a wrong anwswer, and so on. The numbers below each bar indicate explicitly the height of the bar. On panel \textbf{(a)} we consider the whole test set, on panel \textbf{(b)} we only consider the examples which were correctly classified by the arithmetic average EDeiT model {and on panel \textbf{(c)} we only consider the examples which were correctly classified by the geometric average EDeiT model.}}
    \label{fig:www_vit}
\end{figure*}

This results in the ViT ensembles being more often in a (RRR) configuration, as shown in Fig.~\ref{fig:www_vit}--left. However, the higher number of (RRR) is overcompensated by a lower number of (RRW) [and of correctly classified (RWW), Fig.~\ref{fig:www_vit}], which eventually result in the DeiT ensemble having a better performance.

\begin{figure*}[t]
    \centering
    \includegraphics[width=.99\textwidth]{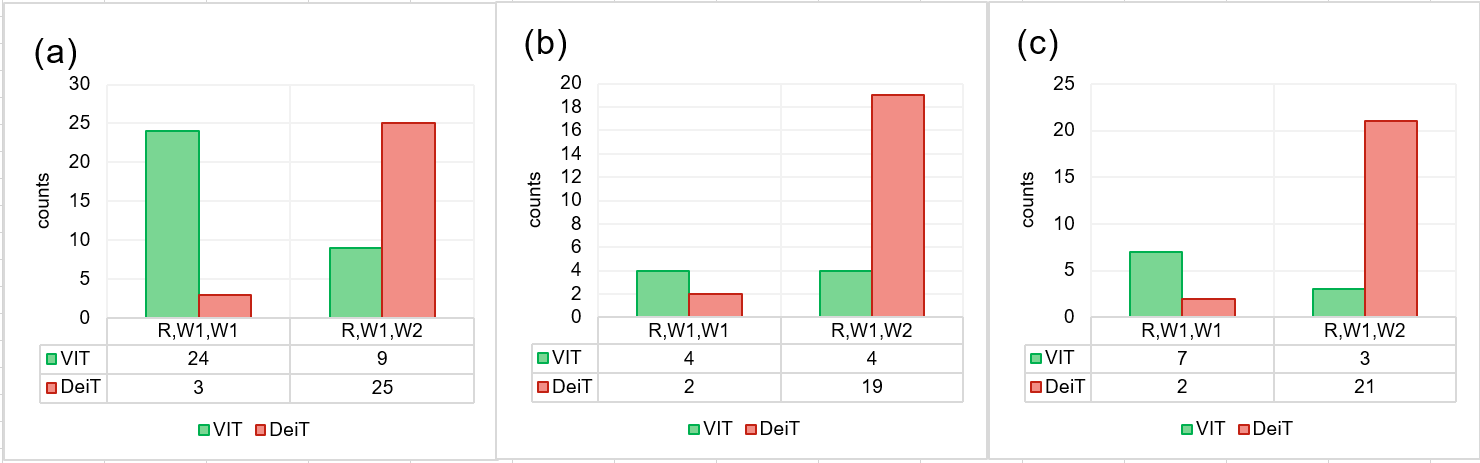}
    \caption{Comparison between ensembled VITs (B16, B32, L32) with ensembled DeiTs. 
    \textbf{(a)}: The bars show how many times, out of the full ZooLake test set, the single learners within the ensemble model gave one correct answer (R) and two wrong answers that were indicating the same class ($W_1$ and $W_1$). The bars indicate one R answer and two wrong answers that differ from each other ($W_1$ and $W_2$).
     \textbf{(b)}: Same as (a), but only restricted to the examples that resulted in a correct classification by the arithmetic average EDeiT model. {\textbf{(c)}: Same as (a), but only restricted to the examples that resulted in a correct classification by the geometric average EDeiT model.}}
    \label{fig:mistake-cases_vit}
\end{figure*}

\begin{figure}[tbh]
\centering
\includegraphics[width=\columnwidth]{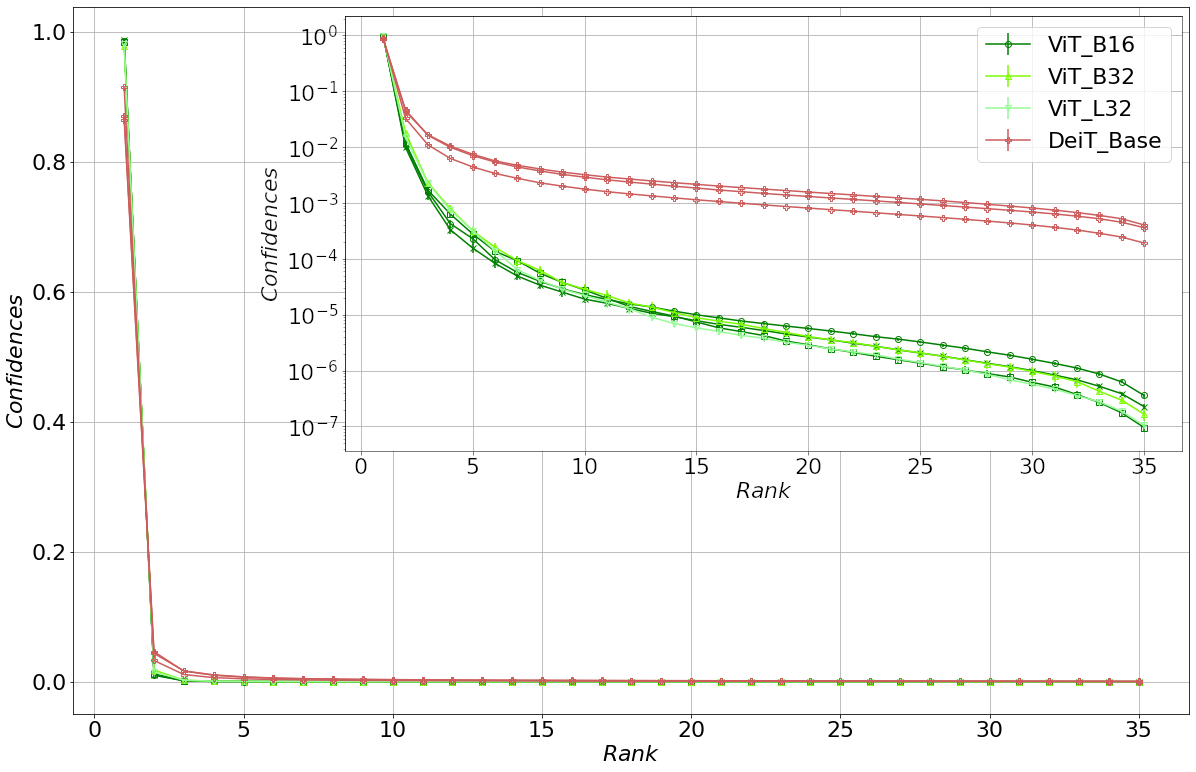}
\caption{For various model types, the mean confidences across all classes (x-axis) are shown. When it comes to classification, the VIT models (greens) are more confident than the DeiT-base models (red), as can be observed in the outset figure (linear scale). The inset figure shows the same in log scale.}
\label{fig:confs_vit}
\end{figure}

This is also seen in the higher rate of correctly classified (RWW) examples of DeiTs with respect to ViTs (Fig.~\ref{fig:mistake-cases_vit}), analogously to what we showed in App.~\ref{app:RW1W2} for CNNs. Analogous considerations also apply here, with the confidence vectors being qualitatively dissimilar from DeiTs (and similar to CNNs) also in the case of ViTs (Fig.~\ref{fig:confs_vit}).

\bibliographystyle{unsrt}
\bibliography{marco}

\end{document}